\pdfoutput=1
\documentclass[a4paper]{article}

\usepackage{INTERSPEECH2019}
\usepackage{amsmath}
\usepackage{hyperref}
\usepackage{multirow}
\usepackage{algorithm}

\newcommand*{\Es}{\operatorname{E_s}}

\newcommand*{\Ec}{\operatorname{E_c}}

\newcommand*{\D}{\operatorname{D}}

\newcommand*{\X}{\mathcal{X}}

\newcommand*{\N}{\mathcal{N}}
\newcommand*{\E}{\mathop{\mathbb{E}}}

\title{One-shot Voice Conversion by Separating Speaker and Content Representations with Instance Normalization}
\name{Ju-chieh Chou, Cheng-chieh Yeh, Hung-yi Lee}
\address{
  College of Electrical Engineering and Computer Science, National Taiwan University}
\email{\{r06922020,r06942067,hungyilee\}@ntu.edu.tw}

\begin{document}

\maketitle
\begin{abstract}
  %Recently, unsupervised voice conversion(VC) have been successfully adapted to multi-target scenario. It means that a single model can be trained to convert the voice to many different speakers. However, such model suffers from the limit that it can only convert the voice to the speaker set in the training data, which narrows the applicable scenario of VC. In this paper, we proposed a one-shot unsupervised VC approach. By separately learning the speaker and content representations from the data, our model is able to perform VC by an examplar which may not appear in the the training data. Also, we don't have to specify any domain information of the each utterance (e.g. speaker, recording condition) in the training process, which make data collection more efficient. In addition to the performance measurement of the conversion, we also demonstrate that this model is able to learn meaningful static representations without any supervision.
 
Recently, voice conversion (VC) without parallel data has been successfully adapted to multi-target scenario in which a single model is trained to convert the input voice to many different speakers. 
However, such model suffers from the limitation that it can only convert the voice to the speakers in the training data, which narrows down the applicable scenario of VC. 
In this paper, we proposed a novel one-shot VC approach which is able to perform VC by only an example utterance from source and target speaker respectively, and the source and target speaker do not even need to be seen during training. 
This is achieved by disentangling speaker and content representations with instance normalization (IN).
Objective and subjective evaluation shows that our model is able to generate the voice similar to target speaker.
In addition to the performance measurement, we also demonstrate that this model is able to learn meaningful speaker representations without any supervision.

\end{abstract}

\section{Introduction}
VC aims to convert the non-linguistic information of the speech signals while maintaining the linguistic content the same. The non-linguistic information may refer to speaker identity~\cite{miyoshi2017voice, nakashika2014voice, kinnunen2017non}, accent or pronunciation~\cite{hojo2018automatic,chen2018generative} to name a few. 
VC can be useful in some down-stream tasks like multi-speaker text-to-speech~\cite{jia2018transfer,tjandra2018machine} and expressive speech synthesis~\cite{skerry2018towards,wang2018style}, and also some applications like speech enhancement~\cite{kumar2016improving,kaneko2017generative,tanaka2018wavecyclegan} or pronunciation correction~\cite{hojo2018automatic}, and so on. 
In this paper, we will focus on the problem of speaker identity conversion. 
% related work

Prior works on VC can be roughly categorized into two types, a supervised one and an unsupervised one. Supervised VC has achieved great performance~\cite{mohammadi2014voice,nakashika2014high,sun2015voice, chen2014voice}. However, it requires frame-level alignment between source and target utterance. If there is a huge gap between source and target domain, inaccurate alignment may hurt the performance of the conversion. More important, collecting parallel data is difficult and time-consuming, which make supervised VC not a desirable framework if we want to have the flexibility of adapting it to some new domains. 

%As a result, 
Unsupervised VC recently became an actively investigated problem due to its efficiency in data collection. It means that we do not have to collect parallel data, but to utilize non-parallel data to train the VC system. Some works try to incorporate ASR system to perform unsupervised VC \cite{yeh2018rhythm,sun2016personalized,xie2016kl}. 
By translating the speech to phoneme posterior sequences, and then synthesizing the speech with the target domain synthesizer, unsupervised VC can be achieved. However, The performances of this kind of approaches highly depend on the accuracy of the ASR system, and will corrupt if the ASR system is not well-functioned. 
Some other works try to utilize deep generative model like VAE~\cite{kingma2013auto} or GAN~\cite{goodfellow2014generative} to do unsupervised VC ~\cite{kaneko2017parallel,kameoka2018stargan,kameoka2018acvae,hsu2017voice,chou2018multi}. These works formulate VC as a domain mapping problem, aiming to learn networks that can transfer utterances among different domains. 
These works are able to generate speech with good quality and can convert the speaker characteristics successfully. However, the major limitation of these works is that they can not synthesize the voice of the speakers who were never seen in training phase.
%yeh% Some "works" try to incorporate ASR system to perform "unsupervised" VC不過前面都說了還是加好了XDnonparallel VC?
%yeh% ... by translating the speech... and then synthesizing ...
%yeh% "the performances" of this kind of "approaches" highly depend on the "accuracy" ... and will corrupt if the ASR system
%yeh% Some other works [cite here] try to utilize ...,

%yeh% cycleGAN cite一下？ 會不會超過(可
%yeh% 不要好了 整句改掉XD?
%yeh% These works formulate VC as a domain mapping problem, aiming to learn networks that can transfer utterances among different domains.
%yeh% These works are able to generate speech with good quality and can convert the speaker characteristics successfully. However, the major limitation of these works is that they can not synthesize the voice of the speakers who were never seen in training phase.

\begin{figure}[t]
  \centering
  \includegraphics[width=\linewidth]{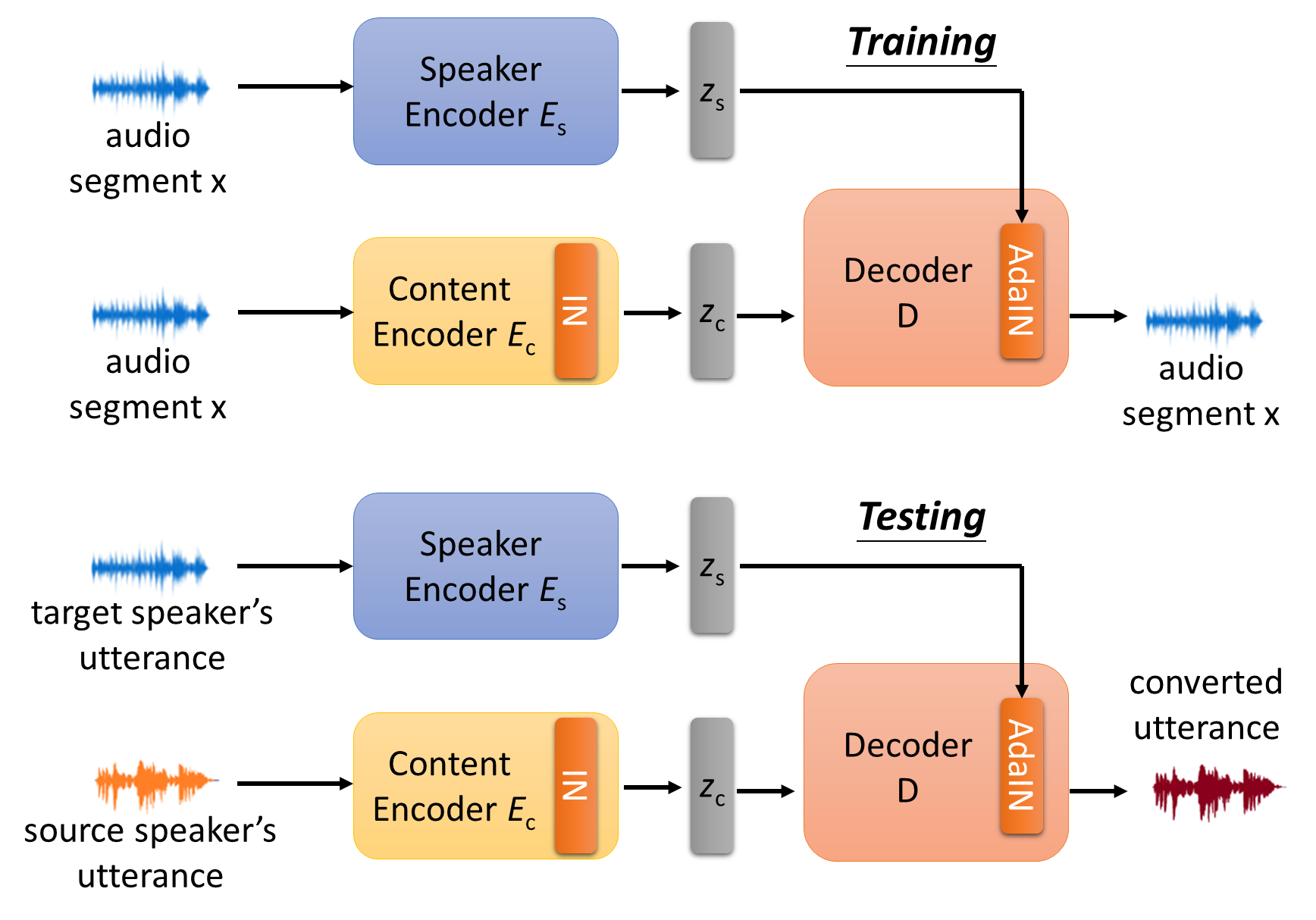}
  \caption{Model overview. $\Es$ is speaker encoder; $\Ec$ is content encoder and $\D$ is decoder. IN is instance normalization layer without affine
transformation. AdaIN represents adaptive instance normalization layer.}
  \label{fig:model}
\end{figure}

%Motivated by the fact that speech signals inherently carry both static information and linguistic information. %這句文法不對
Speech signals inherently carry both static information and linguistic information.
The static part such as speaker, acoustic condition is time-independent and merely changes during the whole utterance, while the linguistic part may change dramatically every several frames. 
Here we   assume   an utterance can be factorized into a speaker representation plus a content representation. 
%To achieve the above-mentioned goal, %which goal? J:disentangle :)
To disentangle speaker and content representation, our model consists of three components: a speaker encoder, a content encoder and a decoder in \autoref{fig:model}. 
The speaker encoder is trained to encode the speaker information into the speaker representation. The content encoder is trained to encode only the linguistic information into the content representation. And then the task of the decoder is to synthesize the voice back by combining these two representations. 
%To disentangle the speaker information and content information, 
We utilize instance normalization~\cite{ulyanov2016instance} without affine transformation in the content encoder to normalize the channel statistics, which control the global information.
In this way, the global information such as speaker information    is removed from the representation encoded by the content encoder.
And also, adaptive instance normalization~\cite{huang2017arbitrary} is utilized in the decoder, the corresponding affine parameters are provided by the speaker encoder. By doing this, the global information needed in the decoder is controlled by the speaker encoder. With the designed architecture, our model is encouraged to learn factorized representations. This kind of factorization enable our model to perform one-shot voice conversion as follows: with one utterance from source speaker and another utterance from target speaker, we first extract the speaker representation from the target utterance, and then extract the content representation from the source utterance, and finally combine them with the decoder to generate the converted result as in \autoref{fig:model}. 
It is worth mentioning that our model does not require any speaker label of the utterances during the training process, which makes the data collection easier. 
Interestingly, the speaker encoder learns a meaningful speaker embeddings even if we do not provide any speaker label.
In terms of applying factorization techniques to speech, some prior works proposed using adversarial training to remove certain attributes from an utterance~\cite{hsu2018disentangling,chou2018multi}. However, with the cost of training an extra discriminator network, lots more computational resources are used. 
%Lee: 如果空間不夠就不要提 computational resources ，因為instability才是比較嚴重的問題
Also, adversarial training suffers from instability problem, which makes the training process difficult. 
In our proposed approach, we simply use the technique of instance normalization instead of adversarial training to remove the speaker information in an utterance, which substantially reduces the computation and makes the training process easier.

%Yeh% In terms of VC with factorization techniques, some prior works proposed using adversarial training[cite] to remove certain attributes from an utterance to perform one-shot VC(??還是VC就好), but with the cost of training an extra discriminator network which requires lots more computational resources. Also, adversarial training suffers from instability problem, [cite一下]
%J,因為有wei-ning那篇就刪掉VC% In terms of VC with factorization techniques, some prior works proposed using adversarial training[cite] to remove certain attributes from an utterance to perform one-shot VC(??還是VC就好), but with the cost of training an extra discriminator network which requires lots more computational resources. Also, adversarial training suffers from instability problem, [cite一下]
%In our proposed method, we simply use the technique of "Instance Normalization" instead of adversarial training to remove the speaker information in an utterance, substantially reduce the computation and make the training process easier.%
%DONE%

%怎麼train的
%Our model is a VAE architecture. The content encoder will encode an acoustic feature segment into a uni-variance Gaussian distribution. Given a sample from this distribution and the speaker representations encoded by the speaker encoder, the decoder will reconstruct the original acoustic feature segment. Also, the KL-divergence between the posterior distribution encoded by content encoder and a prior distribution $\N(0, I)$ will be minimized. It is worth mentioning that our model do not require any speaker label of the utterances during the training process, which make the data collection more efficient. Interestingly, the speaker encoder learns a meaningful speaker embeddings even if we do not provide any speaker label.

Our contribution is three-fold:
\begin{enumerate}
\item Our proposed model is able to do one-shot VC without any supervision.  
\item The efficacy of instance normalization on disentangling representations for VC is verified.
\item We demonstrate that our model is able to learn meaningful speaker embedding as a side effect.
\end{enumerate}

%The training process of this model includes two stages. The first stage we train the VAE part of the model. The content encoder and generator form a VAE-like structure. The second stage, we freeze the parameters of the static and dynamic encoder, and train the generator and discriminator, which form a GAN. 

%This template can be found on the conference website. Templates are provided for Microsoft Word\textregistered, and \LaTeX. However, we highly recommend using \LaTeX when preparing your submission. Information for full paper submission is available on the conference website.

\section{Proposed Approach}

\subsection{Variational autoencoder} \label{subsec:VAE}
Let $x$ be the acoustic feature segment, and $\X$ be the collection of all the acoustic segments in the training data. 
Let $\Es$ be the speaker encoder, $\Ec$ be the content encoder, and $D$ be the decoder. $\Es$ is trained to generate the speaker representation $z_s$. And $\Ec$ is trained to generate content representation $z_c$. 
We assume that  $p(z_c \vert x)$ is a conditionally independent Gaussian distribution with unit variance as in~\cite{liu2017unsupervised}, which means $p(z_c \vert x) = \N(\Ec(x), I)$.
%We use mean absolute error (MAE) as the reconstruction loss.
The reconstruction loss is given as in \autoref{eq1}.
\begin{equation}
\begin{split} 
L_{rec}(\theta_{\Es},\theta_{\Ec}, \theta_{\D})&=\E\limits_{x \sim p(x), z_c\sim p(z_c \vert x) } [\Vert\D(\Es(x), z_c)-x\Vert^1_1].
\end{split}
\label{eq1}
\end{equation}
We uniformly sample an acoustic segment $x$ from $\X$ during training process (that is, $p(x)$ in \autoref{eq1} is an uniform distribution over $\X$).
To match the posterior distribution $p(z_c|x)$ to the prior $N(0, I)$, the KL divergence loss will be minimized.
%yeh% Follow the original VAE work[cite], to match the posterior distribution $p(z_c|x)$ to the prior $N(0, I)$, another KL divergence loss (Eq. 2) need to be minimized, too. Since we further assume the prior Gaussian is with unit variance, this term actually degenerates to a L2 regularization.
Since we assume unit variance, the KL divergence reduces to L2 regularization. The KL divergence term is given as in \autoref{eq2}.
\begin{equation}
L_{kl}(\theta_{\Ec})=\E\limits_{x \sim p(x)} [\Vert {\Ec(x)}^2 \Vert^2_2].
\label{eq2}
\end{equation}
The objective function for VAE training is to minimize the combination of the two terms with weighted hyper-parameters $\lambda_{rec}$ and $\lambda_{kl}$.
\begin{equation}
\min\limits_{\theta_{\Es},\theta_{\Ec}, \theta_{\D}} L(\theta_{\Es},\theta_{\Ec}, \theta_{\D})=\lambda_{rec} L_{rec} + \lambda_{kl} L_{kl} 
\label{eq4}
\end{equation}

\subsection{Instance Normalization for Feature Disentanglement}
%Because there is no significant difference between the two encoders $E_s$ and $E_c$, 
At the first glance, it is unclear how could the two encoders $\Es$ and $\Ec$  encode speaker and content information respectively based on the description in Section~\ref{subsec:VAE}.   
In this paper, we find that simply adding Instance normalization (IN) without affine transformation to $\Ec$   can remove the speaker information while preserving the content information.
Similar idea has been verified to be effective for style transfer in computer vision~\cite{huang2017arbitrary}. 

The formula of instance normalization (IN) without affine transformation is given as in \autoref{eqin}.
%Lee: x 前面用過了，是指 audio segment ，這邊不要再用
Here $M$ is the feature map of the output of the previous convolutional layer, and $M_c$ represents the $c$-th channel, which is a $W$-dimensional array.
%$W$ is the  width of the feature map.
Here each channel is an array instead of a matrix because 1-D convolution is applied rather than 2-D.
To apply IN, we have to compute the mean $\mu_{c}$ and standard variation  $\sigma_{c}$ of the $c$-th channel first.
\begin{equation}
\begin{split}
\mu_{c}  &= \frac{1}{W} \sum^{W}_{w=1} M_c[w], \\
\sigma_{c} &= \sqrt{ \frac{1}{W} \sum^{W}_{w=1} (M_c[w] - \mu_{c})^2 + \epsilon},
\end{split}
\label{eqnorm}
\end{equation}
where $M_c[w]$ is the $w$-th element in $M_c$.
$\epsilon$ in \autoref{eqnorm} is simply a small value to avoid numerical instability.
Then in IN, each element in the array $M_c$ is normalized into $M_c^\prime$ as below. 
\begin{equation}
M^\prime_c[w] = \frac{M_c[w] - \mu_{c}}{\sigma_{c}} \label{eqin}
\end{equation}
The normalized $M_c^\prime$  are processed by the following deep network layers.
We utilize IN layer in content encoder to prevent the content encoder from learning domain information. So as to enforce the model to extract speaker information from speaker encoder and content information from content encoder respectively.

To further enforce the speaker encoder to generate speaker representation, we provide the speaker information to decoder by adaptive instance normalization (adaIN) layer~\cite{huang2017arbitrary}. 
In adaIN layer, the decoder first normalizes the global information by IN, and the speaker encoder provide the  global information.
The formula is given as followed.
\begin{equation}
M^\prime_c[w] =  \gamma_c \frac{M_c[w] - \mu_{c}}{\sigma_{c}}  + \beta_c.
%adaIN(x) = \gamma \frac{x - \mu(x)}{\sigma(x)} + \beta
\label{eqadain}
\end{equation}
$\mu_{c}$ and $\sigma_{c}$ are computed as \autoref{eqnorm}.
$\gamma_c$ and $\beta_c$ for each channel are the linear transformation of the output of speaker encoder $\Es$.

\section{Implementation Details}

% rchitecture, linear increased lambda, instance normalization, mel-spectrograms 
\subsection{Architecture}
\begin{figure}[t]
  \centering
  \includegraphics[width=0.92\linewidth]{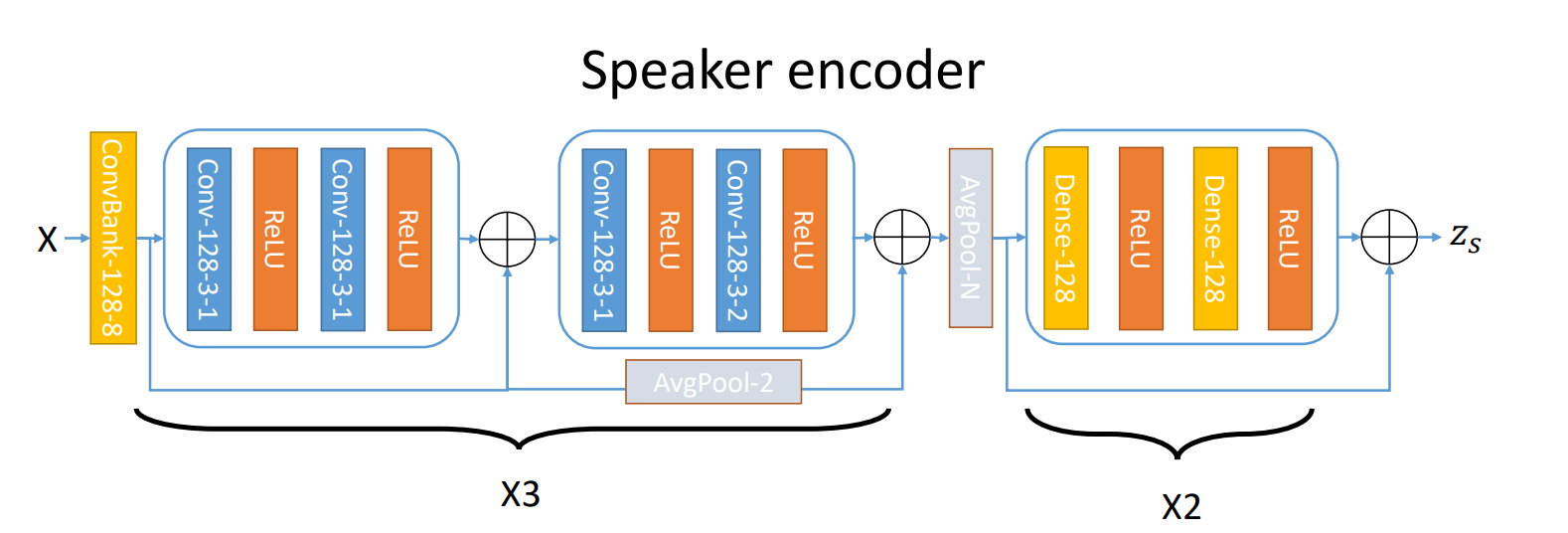}
  \includegraphics[width=0.92\linewidth]{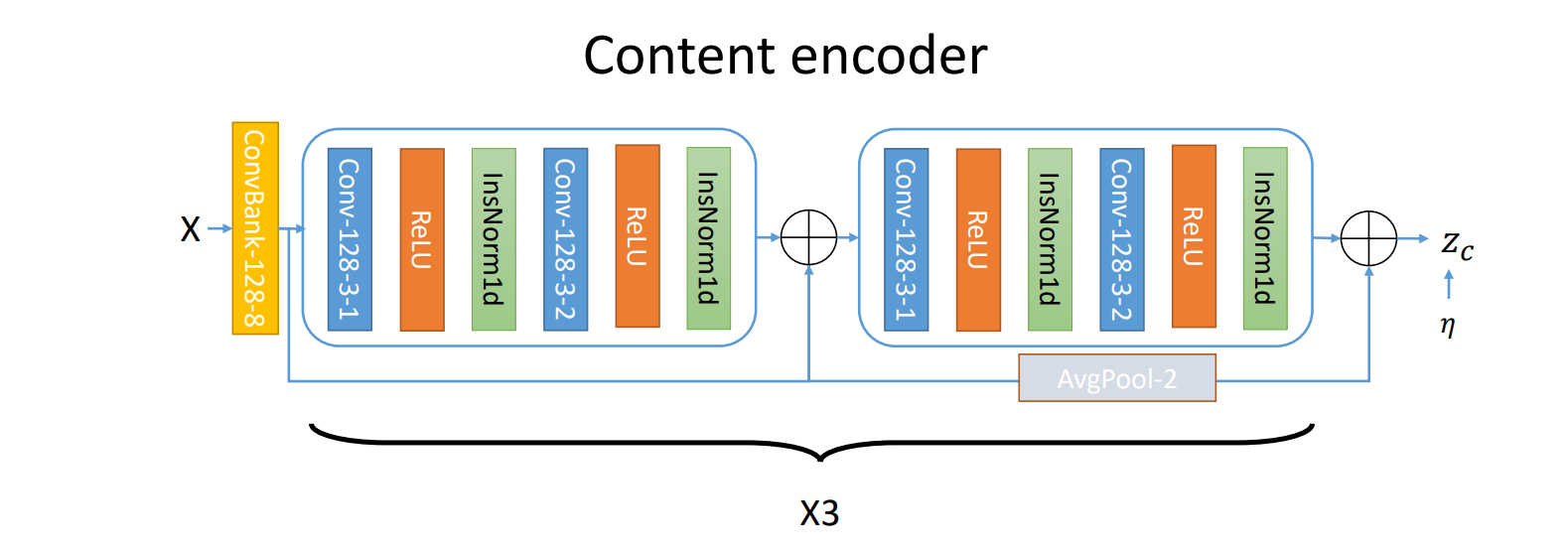}
  \includegraphics[width=0.92\linewidth]{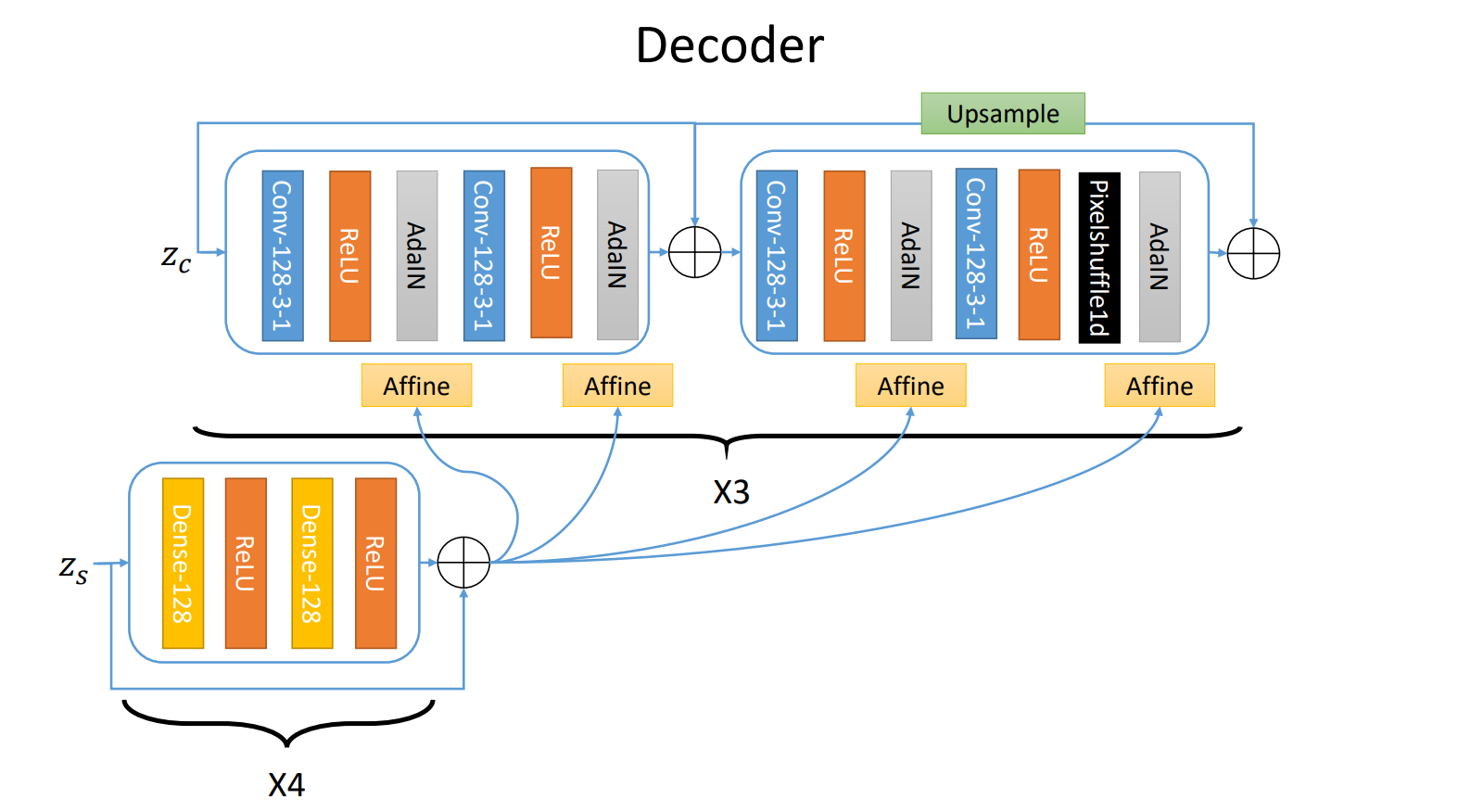}
  \caption{The architecture of the encoders and decoder. }
  %ConvBank-C-K means ConvBank layer with channels equals C and bank size equals K. Conv-C-K-S are convolutional layers with channels equals C, kernel size equals K and stride equals S. }
  \label{fig:e}
\end{figure}

We use Conv1d layers in encoders and decoder to process all the frequency information at a time as in \autoref{fig:e}. The ConvBank layer is used in both the speaker encoder and content encoder to better capture long-term information~\cite{wang2017tacotron}. 
%We use average pooling and nearest-neighbor upsampling for the residual connection if the widths of the two connected layers are different. 
We apply average pooling over time to the speaker encoder so as to enforce the speaker encoder to learn global information only. 
Instance normalization layers are used in content encoder to normalize the global information. PixelShuffle1d~\cite{shi2016real} layer is used in the decoder for upsampling. adaIN layer is used to provide global information to the decoder. The speaker representation $z_s$ is first processed by a residual DNN, and then transformed by an affine layer before get into each adaIN layer. 

\subsection{Acoustic feature}
We use mel-scale spectrograms as the acoustic feature. We first trimmed out the silence and normalize the volume, and then convert the audio to 24kHz.  After that, we perform STFT to the audio with a 50 milliseconds window length, a 12.5 milliseconds hop length, and a 2048 STFT window size. And then transformed magnitude of the spectrograms to 512-bin mel-scale spectrograms. The mel-scale spectrograms are normalized by subtract mean and divide standard deviation.
To convert the mel-scale spectrograms back to waveform, we apply the approximate inverse linear transformation to recover the linear-scale spectrograms~\cite{engel2019gansynth}. And phase is reconstructed by Griffin-Lim algorithm with $100$ iterations.

\subsection{Training details}
We trained the proposed model by ADAM optimizer with a $0.0005$ learning rate, and $\beta_1=0.9$, $\beta_2=0.999$. We set the batch size to $256$. To prevent the model from over-fitting, we apply dropout to each layers with a $0.5$ dropout rate and a $0.0001$ weight decay. 
%ReLU activation are used in all layers except for the last layer of encoders and decoder.
$\lambda_{rec}$ is set to $10$ and $\lambda_{kl}$ is set to $0.01$. We trained the model for $200000$ iterations (mini-batches). Further details may be found in our implementation code: \url{https://github.com/jjery2243542/adaptive_voice_conversion}

\section{Experiments}
We evaluated our model on CSTR VCTK Corpus~\cite{veaux2016superseded}. The audio data were produced by 109 speakers in English with different accents.
%such as English, American, and India. Each speaker uttered different sets of sentences. 
We randomly selected 20 speakers' utterances as our testing set, and the rest utterances will be split to 90\% training set and 10\% validation set. While we set the segment length to be 128 during training, because of the fully-convolutional architecture, the model can process input with any length at inference stage. After removing all the utterances less than 128 frames, the training set contains about 16000 utterances.
%yeh% While we set the segment length to be 128 during training, because of the fully-convolutional architecture, the model can process input with any length at inference stage. After removing all the utterances less "than" 128 frames, the training set contains about 16000 utterances.

\subsection{Evaluation of disentanglement}\label{sec:dis}
%To see the effect of IN, We trained another verification network to verify whether the IN successfully remove the information of speaker characteristic. The network is a 5-layer DNN with 1024 neurons and ReLU activation. The verification network is trained to classify speaker identity given the latent representation encoded by the content encoder. We compared the result between content encoder with IN, content encoder without IN and content encoder without IN plus speaker encoder with IN. The result is given at \autoref{tab:ver}. We can see that the accuracy for speaker identity prediction drop a lot when we apply the IN to content encoder. But due to the fact that the speaker encoder is able to control the channel statistics of decoder by adaIN, the model tends to learn the speaker information from speaker encoder. So the accuracy is low even if we did not apply the IN to content encoder. If we apply IN to speaker encoder, due to the average pooling over time layer in speaker encoder, the speaker encoder will learn nothing. Then the accuracy for speaker identity prediction will increase.
To see the effect of IN layer, we performed an ablation study to verify it could help content encoder remove the information of speaker characteristics. We trained another network (5-layer DNN with 1024 neurons and ReLU activation) to classify speaker identity given the latent representation encoded by the content encoder. 
We compared the classification accuracy under three settings which were "content encoder with IN", "content encoder without IN" and "content encoder without IN while speaker encoder with IN", respectively. The results are shown in \autoref{tab:ver}. %Lee: 為什麼沒有 "content encoder with IN while speaker encoder with IN"?
%因為跟第3個一樣
We can see that the classification accuracy is apparently lower when IN is applied to the content encoder.
But we also found the accuracy was not as high as expected even if we did not apply IN to the content encoder. 
This was probably because by the fact that the speaker encoder is able to control the channel statistics of decoder by adaIN, the whole model tends to learn speaker information from the speaker encoder rather than from the content encoder. 
To further confirm this assumption, we tested the classification accuracy under the third settings mentioned above, which was not to apply IN to the content encoder but applied it to the speaker encoder. As we can see, due to the average pooling over time property combined with IN layer (output zero-vector), the speaker encoder could no longer possess the complete speaker information, thus the whole model tended to "flow" more speaker information through content encoder, increasing the classification accuracy.

\begin{table}[th]
  \caption{The accuracy for speaker identity prediction on content representation. Smaller value means less speaker information in the content representation.}
  \label{tab:ver}
  \centering
  \begin{tabular}{ccc}
    \toprule
    \multicolumn{1}{c}{\textbf{$\Ec$ with IN}} & 
    \multicolumn{1}{c}{\textbf{$\Ec$  w/o IN}} & 
    \multicolumn{1}{c}{\textbf{$\Ec$  w/o IN + $\Es$  with IN}} \\
    \midrule
     $0.375$  & $0.658$    & $0.746$         \\
    \bottomrule
  \end{tabular}
\end{table}

\subsection{Speaker embedding visualization}
\begin{figure}[t]
  \centering
  \includegraphics[width=\linewidth]{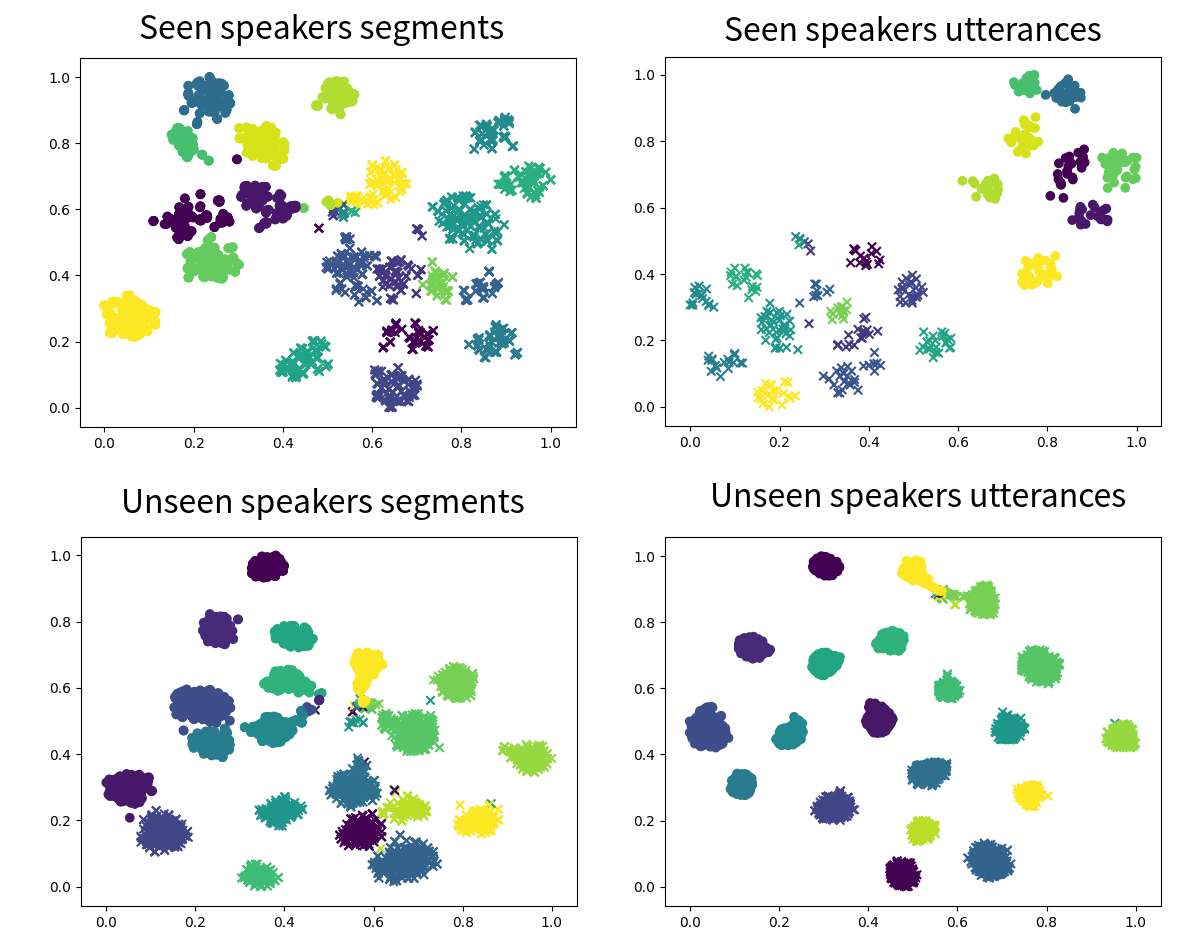}
  \caption{The visualization of speaker embedding. 'x' are female speakers and 'o' are male speakers. Segments are randomly sampled from validation set and testing set. }
  \label{fig:emb}
\end{figure}

We found that the speaker encoder learned meaningful embeddings related to speakers even if we did not explicitly add any objective or constraint to the encoder~\cite{ravanelli2018learning}. We inputted both seen and unseen (during training) speakers' utterances through speaker encoder and plotted their embeddings in 2D space with t-SNE in \autoref{fig:emb}. We found that utterances spoken by different speakers were well-separated. We also conducted experiments on classifying speaker id with these embeddings. The setting was the same as \autoref{sec:dis}. Seen speakers achieved $0.9973$ accuracy and unseen speakers achieved $0.9998$ accuracy, indicating that the speaker encoder learned reasonable representations in the embedding space.
%yeh% meaningful embeddings "related to speakers" ... add any objective or constraint to "the encoder". We inputted both seen and unseen (during training) speakers' utterances through speaker encoder and plotted their embeddings in 2D space with t-SNE in Fig. 3. We found that utterances spoken by different "speakers were" well-separated. We also conducted experiments on classifying speaker id with "these embeddings". The setting "was" the same as subsection3.1. ..."achieved"..."achieved".

\subsection{Objective evaluation}

\subsubsection{Global variance}
\begin{figure}[t]
  \centering
  \includegraphics[width=\linewidth]{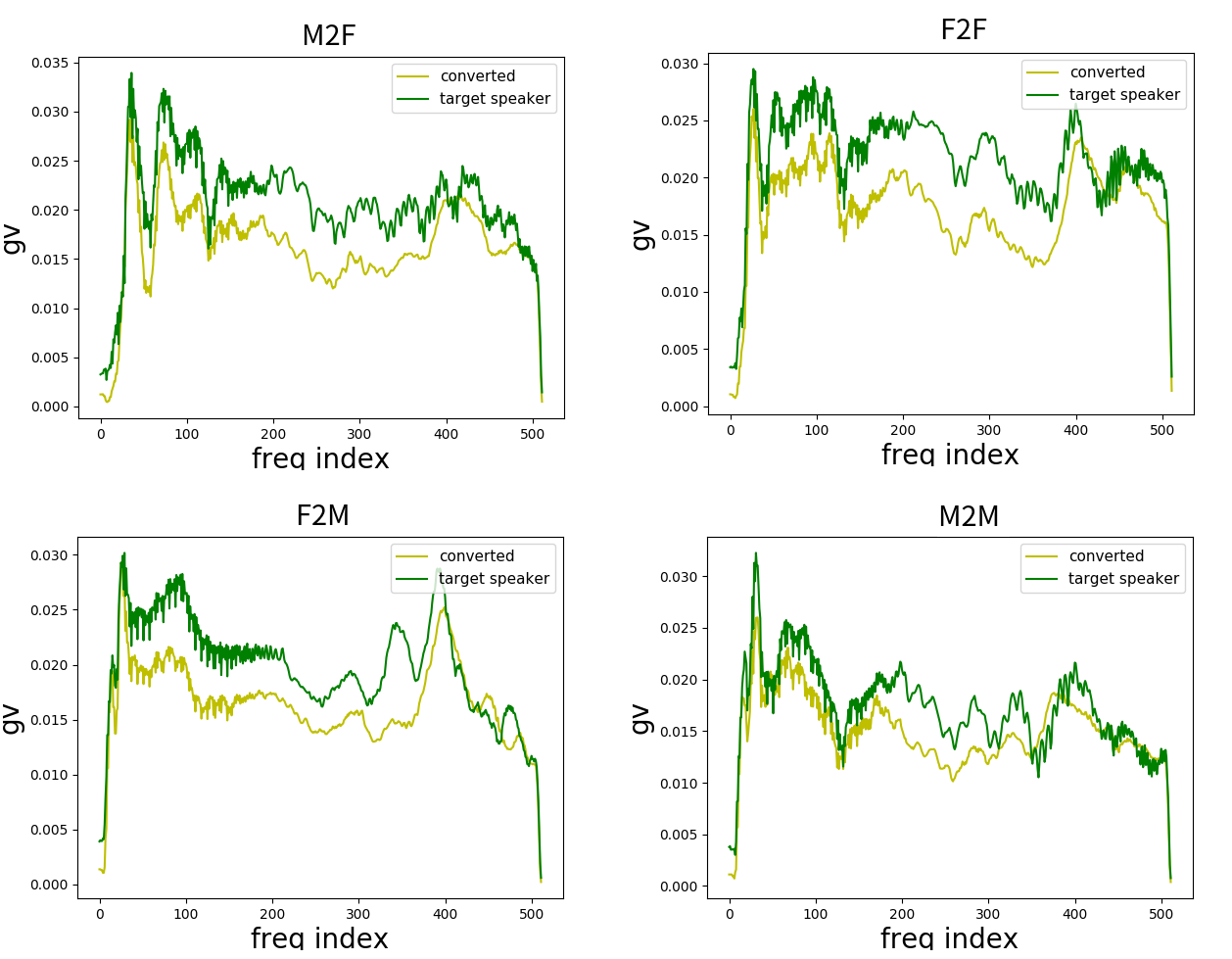}
  \caption{The variance distribution of converted result and target speaker utterances. 100 randomly chosen utterances and converted result are used to calculate the variance.}
  \label{fig:gv}
\end{figure}

To show that our model is able to convert speaker characteristic, we used the global variance (GV) as the visualization of spectral distribution.
Global variance has been used as a way to see whether voice conversion result match to the target speaker in terms of variance distribution~\cite{toda2007speech}. 
We evaluated the global variance for each of the frequency index for 4 conversion examples: male to male, male to female, female to male, and female to female. 
The results are shown in \autoref{fig:gv}, and we found that our generated samples did match to the target speaker in terms of variance distribution.

\subsubsection{Spectrograms example}
\begin{figure}[t]
  \centering
  \includegraphics[width=\linewidth]{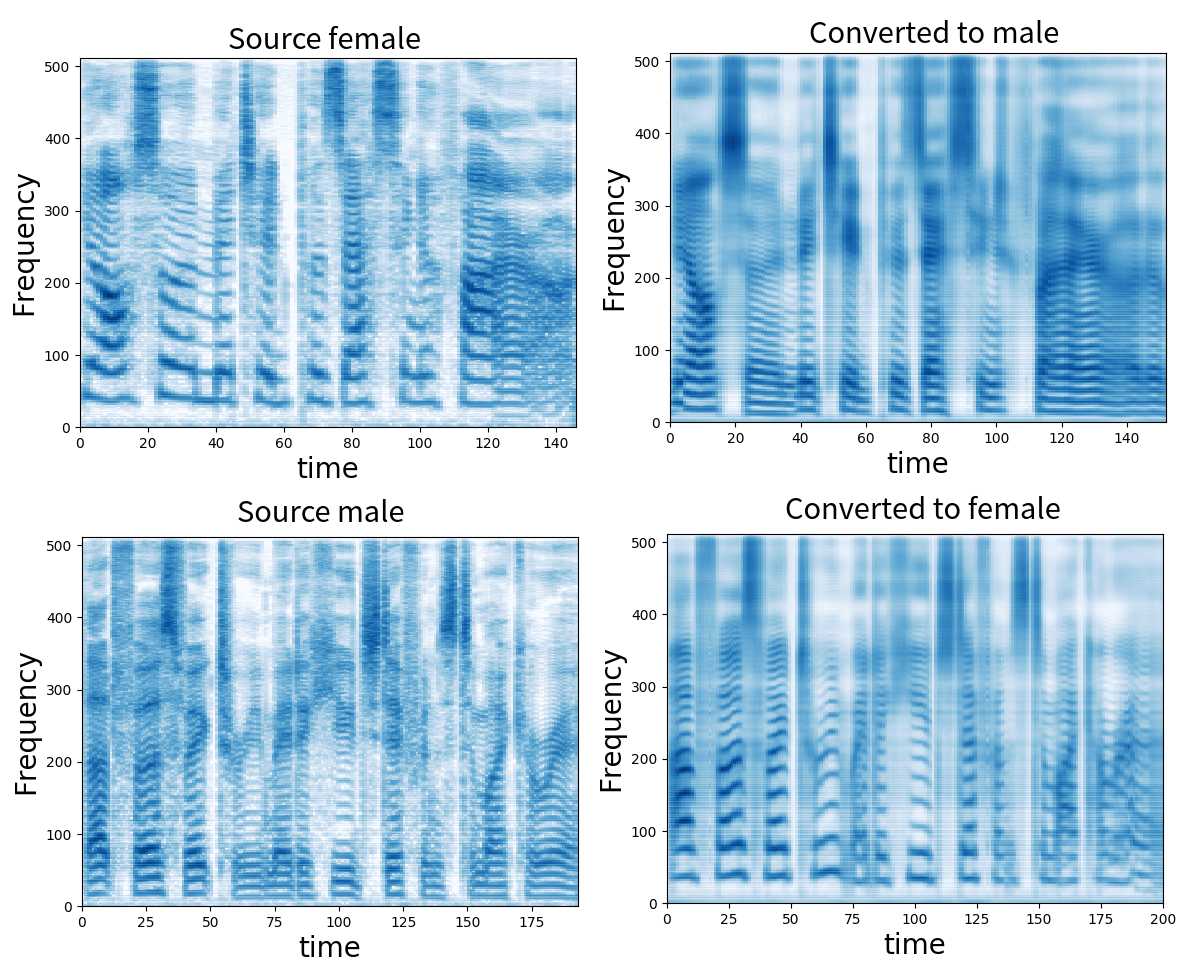}
  \caption{The heatmaps of the spectrogram: the upper left is an utterance spoken by a female speaker. The upper right is the converted result to a male speaker. The lower left is an utterance spoken by a male speaker. The lower right is the converted result to a female speaker.}
  \label{fig:spec}
\end{figure}

Some examples of spectrogram heatmaps are shown in \autoref{fig:spec}. We can see that our model is able to transform the fundamental frequency (f0) and keep the original phonetic content in both male to female conversion and female to male conversion.
%yeh% Some example of "spectrogram" heatmaps (有人這樣說嗎XD?) are shown in Fig. 5.

\subsection{Subjective evaluation}
\begin{figure}[t]
  \centering
  \includegraphics[width=\linewidth]{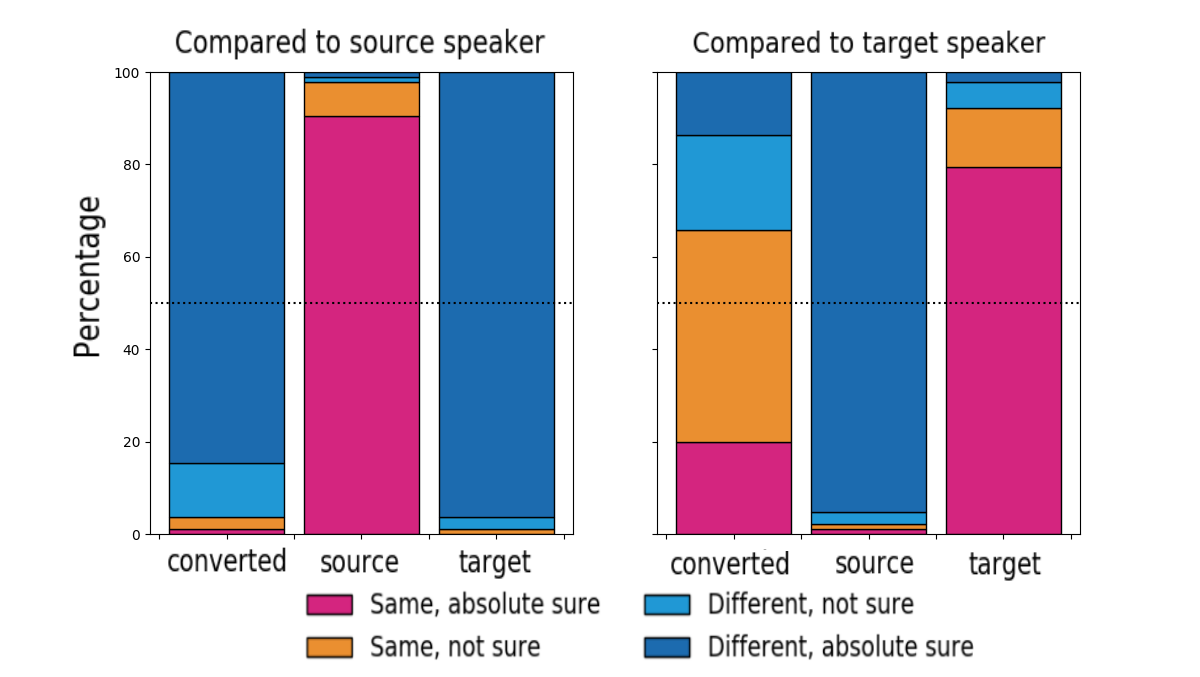}
  \caption{Similarity test. The left one is the comparison to source speaker's utterance. The right one is the comparison to target speaker's utterance.}
  \label{fig:vc}
\end{figure}

Subjective evaluation was performed on converted voice (including male to male, male to female, female to male and female to female, in total four pairs of speakers). The speakers of these four pairs were all unseen during training time, so the converted result of each pair was outputted from our proposed approach by using only one source utterance and one target utterance. 
We then asked the human participants to evaluate the similarity between two utterances with a 4-scale score indicating same absolutely sure, same not sure, different not sure, and different absolutely sure. 
The two utterances were one converted result with either one source speaker utterance or one target speaker utterance. The results are in \autoref{fig:vc}. Our model is able to generate the voice similar to target speaker's. The demo can be found at \url{https://jjery2243542.github.io/one-shot-vc-demo/}.
%yeh% Subjective evaluation was performed on converted voice (including male to male, male to female, female to male and female to female, in total four pairs of speakers). The speakers of these four pairs were all unseen during training time, so the converted result of each pair was outputted from our proposed approach by using only one source utterance and one target utterance. We then ask the human participants to evaluate the similarity between two utterances with a 4-scale score indicating same absolutely sure, same not sure, different not sure, and different absolutely sure. The two utterances were one converted result with either one source speaker utterance or one target speaker utterance. The results are in Fig. 6. Our model is able to generate the voice similar to target speaker's. The demo can be found at ~.

\section{Conclusion}
%We proposed a model to do one-shot VC. The model is trained without any supervision. By properly applying instance normalization, the model will learn factorized representations. The subjective evaluation shows that our model is able to generate the voice similar to target speaker's.
We proposed a novel approach to tackle one-shot unsupervised VC by applying instance normalization to enforce the model to learn factorized representations. 
In this way, we can perform VC to unseen speakers with only one utterance. %needed.
Subjective and objective evaluations showed good result in terms of similarity to target speakers. 
And also, the disentanglement  experiments and visualization showed that in our proposed approach, the speaker encoder  learns a meaningful embedding space without any supervision.

\section{Acknowledgement}
This work is supported by Nvidia.

\end{document}